\documentclass[twocolumn, 9pt]{article}

\usepackage[margin=0.8in]{geometry}
\usepackage{times}
\usepackage{graphicx}
\usepackage{caption}
\usepackage{subcaption}
\usepackage{titlesec}
\usepackage{booktabs}
\usepackage{makecell}
\usepackage{multirow}
\usepackage{tabularx}
\usepackage{ragged2e}
\usepackage{float}
\usepackage{microtype}
\usepackage{upgreek}
\usepackage{changepage}
\usepackage{amsmath, amssymb, amsthm}
\newlength{\extralength}
\newlength{\fulllength}
\usepackage{authblk}
\usepackage{cite}
\usepackage{url}
\usepackage{hyperref}
\hypersetup{
	colorlinks=true,
	linkcolor=blue,
	filecolor=magenta,
	urlcolor=cyan,
	citecolor=blue,
}
\usepackage[capitalize,noabbrev]{cleveref}
\theoremstyle{plain}

\theoremstyle{definition}

\theoremstyle{remark}

\titleformat{\section}
{\large\bfseries}
{\thesection}
{0.5em}
{}
\titleformat{\subsection}
{\normalsize\bfseries}
{\thesubsection}
{0.5em}
{}
\titleformat{\subsubsection}
{\normalsize\itshape}
{\thesubsubsection}
{0.5em}
{}
\titlespacing*{\section}{0pt}{2.5ex}{1.5ex}
\titlespacing*{\subsection}{0pt}{2ex}{1ex}
\titlespacing*{\subsubsection}{0pt}{1.5ex}{0.5ex}

\title{\textbf{Diffusion Models at the Drug Discovery Frontier: A Review on Generating Small Molecules versus Therapeutic Peptides}}

\author[1,2]{Yiquan Wang\thanks{These authors contributed equally to this work.}}
\author[1]{Yahui Ma\protect\footnotemark[1]}
\author[2]{Yuhan Chang}
\author[2]{Jiayao Yan}
\author[2]{Jialin Zhang}
\author[1]{Minnuo Cai}
\author[1]{Kai Wei\thanks{Corresponding author: \href{mailto:kaiwei@xju.edu.cn}{kaiwei@xju.edu.cn}}}

\affil[1]{Xinjiang Key Laboratory of Biological Resources and Genetic Engineering, College of Life Science and Technology, Xinjiang University, Urumqi, Xinjiang, China}
\affil[2]{School of Mathematics and System Sciences, Xinjiang University, Urumqi, Xinjiang, China}

\date{}

\begin{document}
	
	\maketitle
	
	\begin{abstract}
		Diffusion models have emerged as a leading framework in generative modeling, poised to transform the traditionally slow and costly process of drug discovery. This review provides a systematic comparison of their application in designing two principal therapeutic modalities: small molecules and therapeutic peptides. We dissect how the unified framework of iterative denoising is adapted to the distinct molecular representations, chemical spaces, and design objectives of each modality. For small molecules, these models excel at structure-based design, generating novel, pocket-fitting ligands with desired physicochemical properties, yet face the critical hurdle of ensuring chemical synthesizability. Conversely, for therapeutic peptides, the focus shifts to generating functional sequences and designing de novo structures, where the primary challenges are achieving biological stability against proteolysis, ensuring proper folding, and minimizing immunogenicity. Despite these distinct challenges, both domains face shared hurdles: the scarcity of high-quality experimental data, the reliance on inaccurate scoring functions for validation, and the crucial need for experimental validation. We conclude that the full potential of diffusion models will be unlocked by bridging these modality-specific gaps and integrating them into automated, closed-loop Design-Build-Test-Learn (DBTL) platforms, thereby shifting the paradigm from mere chemical exploration to the on-demand engineering of novel~therapeutics.
	\end{abstract}
	
	\vspace{1em}
	
	\subsection{The Bottleneck of Drug Discovery and the Rise of Generative~AI}
	Traditional drug discovery pipelines, reliant on high-throughput screening, which involves the automated testing of large numbers of compounds, and~combinatorial chemistry, a~method for rapidly creating vast libraries of molecules, are characterized by prolonged development timelines, high attrition rates, and~enormous costs. The~entire process from target identification to market approval typically spans 10--15 years~\cite{Singh2023}, with~the clinical development phase alone requiring a median of 8.3 years~\cite{Brown2022}. Despite decades of optimization, clinical success rates remain discouragingly low, with~only approximately 7.9\% of drug candidates entering Phase I trials ultimately receiving regulatory approval~\cite{Kim2023}, though~these rates vary significantly across therapeutic areas and have shown dynamic fluctuations throughout the 21st century~\cite{Zhou2024,Smietana2016,Mullard2016}. Recent advances in cell and gene therapies have demonstrated distinct success rate profiles, offering new prospects for durable treatments~\cite{Phares2025}. The~financial burden is staggering: while historical estimates reached \$2.6~billion per approved drug~\cite{Mullard2014}, more recent analyses suggest mean development costs of approximately \$879 million based on 2000--2018 data~\cite{Sertkaya2024}, though~costs continue to escalate with increasingly complex trial designs and regulatory requirements as evidenced by record-breaking FDA approval trends~\cite{Senior2024}.
	
	The vast chemical space, estimated to contain $10^{60}$ drug-like molecules~\cite{Bohacek1996,Orsi2025}, remains largely unexplored through conventional screening approaches. This estimation, originally derived from molecules up to 30 heavy atoms constructed from organic elements~\cite{Bohacek1996}, has been supported by systematic enumeration studies such as the GDB-17 database containing 166 billion molecules~\cite{Ruddigkeit2012,Reymond2012} and recent explorations of peptide/peptoid chemical space~\cite{Orsi2025}. More conservative estimates suggest approximately $10^{33}$ molecules strictly adhering to Lipinski's rule of five, a~set of physicochemical guidelines used to predict a compound's potential for oral bioavailability~\cite{Polishchuk2013}, yet even this reduced scope represents a vast and largely unsampled space. Generative Artificial Intelligence (AI) offers a paradigm shift, moving from merely screening existing compounds to creating entirely novel molecules tailored to specific needs. This promise is not merely theoretical; the broader field of generative AI has already begun to deliver tangible results, with~dozens of AI-designed small molecules advancing into human clinical trials and demonstrating the potential to significantly shorten discovery timelines and improve success rates~\cite{Jayatunga2024How, Arnold2023Inside, Kanakia2025AI}. Early generative models like Variational Autoencoders (VAEs) \cite{kingma2013auto}, Generative Adversarial Networks (GANs) \cite{goodfellow2020generative}, and~Flow-based models~\cite{dinh2016density,kingma2016improved,papamakarios2017masked} laid the groundwork but faced limitations in generation quality, training stability, and~mode collapse issues. VAEs often produced blurry outputs due to the trade-off between reconstruction and latent loss, while GANs were susceptible to training instability and mode collapse, challenges extensively reviewed in the literature focused on adversarial networks~\cite{sharma2024generative}. Flow-based models, in~turn, encountered computational efficiency limitations. The~distinct trade-offs in performance, stability, and~computational cost across these generative families have been systematically compared in several surveys~\cite{bondtaylor2021deep,vivekananthan2024comparative}.
	
	\subsection{The Emergence of Diffusion~Models}
	Diffusion models have recently emerged as a highly successful framework in generative modeling, demonstrating competitive and robust capabilities in generating high-quality, diverse samples compared to previous approaches~\cite{ho2020denoising}. Their core idea involves a two-step process: a forward diffusion process that incrementally adds Gaussian noise to data according to a predefined variance schedule until it becomes pure noise, and~a reverse denoising process where a trained neural network learns to iteratively denoise samples, effectively generating new data from random noise~\cite{sohl-dickstein2015deep,ho2020denoising}.
	
	The success of diffusion models extends far beyond a single domain. They have achieved revolutionary breakthroughs in fields like computer vision (e.g., DALL-E 2, Stable Diffusion), audio synthesis, and~natural language processing~\cite{ramesh2022hierarchical, rombach2022high, kong2020diffwave, li2022diffusion}, proving an exceptional ability to learn and generate high-quality samples from complex, high-dimensional data. This cross-domain success underscores the framework's inherent flexibility, which makes it particularly attractive for molecular design, where data is inherently multimodal—encompassing continuous 3D coordinates, discrete atom types, graph structures, and~sequential patterns. Moreover, key techniques pioneered for image generation, such as classifier-free guidance~\cite{ho2022classifier} for precise control and latent diffusion~\cite{rombach2022high} for computational efficiency, have been successfully adapted to molecular design challenges~\cite{weiss2023guided, alakhdar2024diffusion}. This powerful combination of generative fidelity and adaptability provides a strong foundation for using diffusion models to create diverse, valid, and~novel therapeutics with desired~properties.
	
	\subsection{Scope and Structure of This~Review}\label{sec:scope}
	
	This review focuses specifically on the recent surge of diffusion models in drug discovery, primarily drawing from the rapidly evolving literature. For~the first time, we systematically compare the application, challenges, and~future prospects of this technology in designing two critical drug modalities: small molecules and therapeutic peptides. These modalities were chosen for their immense clinical and commercial importance and their complementary strengths and weaknesses, which create distinct design challenges perfect for a comparative~analysis.
	
	Small molecules constitute a substantial portion of approved drugs. Recent FDA approval data from 2023 to 2024 indicate that small molecule drugs (new molecular entities, NMEs) accounted for approximately 55--69\% of novel therapeutic approvals~\cite{bai2024comprehensive,wang2025game,mullard2024fda,martins2024food}. In~2023, the~FDA approved 55 new medications consisting of 17 biologics license applications and 38 NMEs, with~small molecules representing approximately 55\% of total approvals~\cite{bai2024comprehensive}. This approval trend continued in 2024, with~50 NMEs approved, further demonstrating the continued importance of small molecule drugs in modern therapeutics~\cite{wang2025game}. Small molecules typically have molecular weights below 900~Da, are orally bioavailable, can penetrate cells to target intracellular proteins, and~are relatively cost-effective to manufacture. They have been successfully applied to a wide range of diseases, from~infectious diseases (antibiotics, antivirals) to chronic conditions (cardiovascular drugs, diabetes medications) to cancer (kinase inhibitors, chemotherapeutics). However, small molecules face significant limitations in targeting certain ``undruggable'' proteins---targets historically considered intractable for small-molecule intervention due to features like lacking well-defined binding pockets or those involving protein-protein interactions with large, flat interfaces~\cite{xie2023recent,nada2024new,xu2025fragment}. These challenging targets have spurred interest in alternative modalities and advanced drug design~approaches.
	
	Therapeutic peptides, by~contrast, represent a rapidly growing class of drugs, with~over 80 peptide drugs currently approved and more than 150 in clinical development~\cite{xiao2025advance}. The~peptide therapeutics field has experienced remarkable growth, driven by advances in peptide chemistry, delivery technologies, and~the clinical success of peptide-based therapeutics such as GLP-1 receptor agonists for diabetes and obesity~\cite{xiao2025advance}. Peptides offer several advantages: high specificity and potency (often binding targets with nanomolar to picomolar affinities), low toxicity (due to degradation into natural amino acids), and~the ability to target protein-protein interactions and extracellular targets that are challenging for small molecules~\cite{xiao2025advance,nada2024new}. These characteristics make peptides particularly valuable for addressing targets previously considered "undruggable" by traditional small molecule approaches~\cite{xie2023recent}. However, peptides face significant biological hurdles, such as poor metabolic stability and potential immunogenicity, which limit their therapeutic application and necessitate specialized design considerations~\cite{Baral2025,Mehrdadi2023,Lamers2022,Verma2021,xiao2025advance}. These complementary strengths and weaknesses make small molecules and peptides ideal for comparative analysis in the context of AI-driven~design.
	
	This review is organized to first introduce the unified framework of diffusion models for molecular generation (Section~\ref{sec:core_engine}). We then dedicate separate sections to their application in designing small molecules (Section~\ref{sec:small_molecules}) and therapeutic peptides (Section~\ref{sec:peptides}), highlighting representative models, performance benchmarks, and~domain-specific challenges. Finally, drawing these threads together, we provide a comprehensive head-to-head comparison, discuss the shared hurdles that transcend modality, and~outline future research directions toward a fully integrated, closed-loop discovery paradigm (Section~\ref{sec:comparison_future}).
	
	\section{The Core Engine: Diffusion Models for Molecular~Generation}\label{sec:core_engine}
	\unskip
	
	\subsection{Representing Molecules for~Diffusion}
	The choice of molecular representation is fundamental to the design of the diffusion process, as~it dictates both the mathematical formulation of the noise process and the architecture of the denoising network~\cite{hu2025target, chen2025multiscale}. For~small molecules, representations primarily fall into two categories. One approach utilizes graph-based representations, where molecules are encoded as graphs with atoms as nodes and bonds as edges~\cite{vignac2022digress, bian2024hierarchical, liu2025graph}, allowing diffusion to operate on features like discrete atom types or continuous latent embeddings~\cite{chen2025multiscale}. An~alternative and increasingly prevalent approach employs 3D coordinate-based representations, treating molecules as point clouds of atomic positions in Euclidean space~\cite{morehead2024geometry, zhang2024equivariant, liu2025clifford}. This latter representation is particularly suited for structure-based drug design, as~it naturally captures spatial relationships critical for protein-ligand interactions and necessitates the use of E(3) equivariant neural networks, architectures designed to respect the natural rotational and translational symmetries of 3D molecules, to~handle rotational and translational symmetries~\cite{satorras2021en, wang2024enhancing, soleymani2024structure, guan20233d, guo2025frame}.
	
	In contrast, the~representation of peptides is shaped by their polymeric nature. The~most straightforward method is sequence-based, encoding peptides as discrete sequences of amino acid tokens, which requires specialized discrete diffusion processes~\cite{austin2021structured, alamdari2023protein, lisanza2025multistate, bai2025mask}. Complementing this, structure-based representations capture the peptide's three-dimensional conformation through the coordinates of backbone and side-chain atoms~\cite{watson2023denovo,chen2025peptide}, or~alternatively, through internal coordinates like torsion angles that inherently respect geometric constraints~\cite{wu2024protein}. These distinct representational paradigms for small molecules and peptides shape the subsequent design of the diffusion models and the type of conditioning information that can be effectively integrated~\cite{li2025thermodynamics, cremer2024latent}.
	
	\subsection{The Mathematics of Diffusion: Forward and Reverse~Processes}
	The diffusion process consists of two Markov chains~\cite{sohl-dickstein2015deep,ho2020denoising}. The~forward process gradually corrupts data $x_0$ by adding Gaussian noise over $T$ timesteps: $q(x_t|x_{t-1}) = \mathcal{N}(x_t; \sqrt{1-\beta_t}x_{t-1}, \beta_t I)$, where $\beta_t$ is a variance schedule. A~key property is that we can sample $x_t$ directly from $x_0$: $q(x_t|x_0) = \mathcal{N}(x_t; \sqrt{\bar{\alpha}_t}x_0, (1-\bar{\alpha}_t)I)$, where $\bar{\alpha}_t = \prod_{s=1}^{t}(1-\beta_s)$~\cite{ho2020denoising}. Here, $\beta_t$ is a small constant from a predefined variance schedule, controlling the amount of noise added at each step. By~defining $\alpha_t = 1 - \beta_t$, the~term $\bar{\alpha}_t = \prod_{s=1}^{t}\alpha_s$ becomes a cumulative product that governs how much of the original signal $x_0$ is preserved at timestep $t$. As~$t$ increases, $\bar{\alpha}_t$ decreases towards zero, signifying that the signal progressively fades into noise. Thus, $\bar{\alpha}_t$ can be intuitively understood as a measure of the signal-to-noise ratio at any given step. The~reverse process learns to denoise: $p_\theta(x_{t-1}|x_t) = \mathcal{N}(x_{t-1}; \mu_\theta(x_t, t), \Sigma_\theta(x_t, t))$. The~model is trained to predict either the noise $\epsilon$ added at each step or the denoised data $x_0$, by~minimizing a variational lower bound on the log-likelihood~\cite{sohl-dickstein2015deep,ho2020denoising}. For~molecular generation, this framework is adapted to handle both continuous (coordinates) and discrete (atom/bond types, amino acid sequences) variables~\cite{austin2021structured,hoogeboom2022equivariant,xu2022geodiff}, often requiring specialized noise processes and network~architectures.
	
	\subsection{Conditional Generation: From Noise to~Purpose}
	Unconditional generation has limited utility in drug design. The~key is \textit{conditional generation}, which steers the generative process toward specific objectives by injecting information—such as a target protein's binding pocket geometry or desired physicochemical properties—into the denoising network at each timestep. Early approaches relied on classifier guidance, which uses a separately trained classifier to steer sampling by adding its gradient to the score function~\cite{dhariwal2021diffusion}. However, a~more recent and popular strategy is classifier-free guidance, which elegantly avoids the need for a separate model by training a single conditional network that can operate both with and without conditioning information, allowing guidance strength to be tuned at inference time~\cite{ho2022classifier}. Another powerful technique, particularly for structure-based tasks, involves integrating conditioning information via cross-attention mechanisms within the denoising network, enabling the model to dynamically attend to relevant features of the conditioning input at each generation step~\cite{huang2024dual}. These techniques provide precise control over the generation process, making them highly suitable for the multi-objective optimization challenges inherent in drug design~\cite{alakhdar2024diffusion,li2025thermodynamics}.
	
	\subsection{Comparison with Other Generative~Approaches}
	
	To appreciate the advantages of diffusion models, it is instructive to compare them with the generative paradigms that were foundational and previously represented the state-of-the-art in drug design, such as Variational Autoencoders (VAEs), Generative Adversarial Networks (GANs), Flow-based models, and~Autoregressive models. This comparison primarily focuses on the overarching generative framework rather than the underlying neural network architecture, as~implementations of these paradigms often share powerful backbones like Graph Neural Networks (GNNs) or Transformers. The~key distinctions, therefore, lie in their training objectives and generation mechanisms. For~example, VAEs like the Junction Tree VAE~\cite{jin2018junction} learn a continuous latent space but often struggle with posterior collapse and may generate chemically invalid structures~\cite{ochiai2023variational, tevosyan2022improving, praljak2023protwave}. Similarly, GANs like MolGAN~\cite{decao2018molgan} can produce diverse molecules but are notoriously difficult to train~\cite{saad2024survey}, frequently suffering from mode collapse and instability~\cite{barsha2025depth, wang2018graphgan}. Flow-based models such as MoFlow~\cite{zang2020moflow}, constrained by their invertible architecture, can lack expressiveness for complex molecular graphs~\cite{madhawa2019graphnvp, mercado2020practical}. Autoregressive models like GraphAF~\cite{shi2020graphaf} can be slow and suffer from error propagation, where an early mistake compromises the entire structure~\cite{segler2018generating, gupta2018generative, alakhdar2024diffusion, wang2025learning, he2019exposure, wang2022your}.
	
\begin{figure*}[H]
		\centering
		\includegraphics[width=1\textwidth]{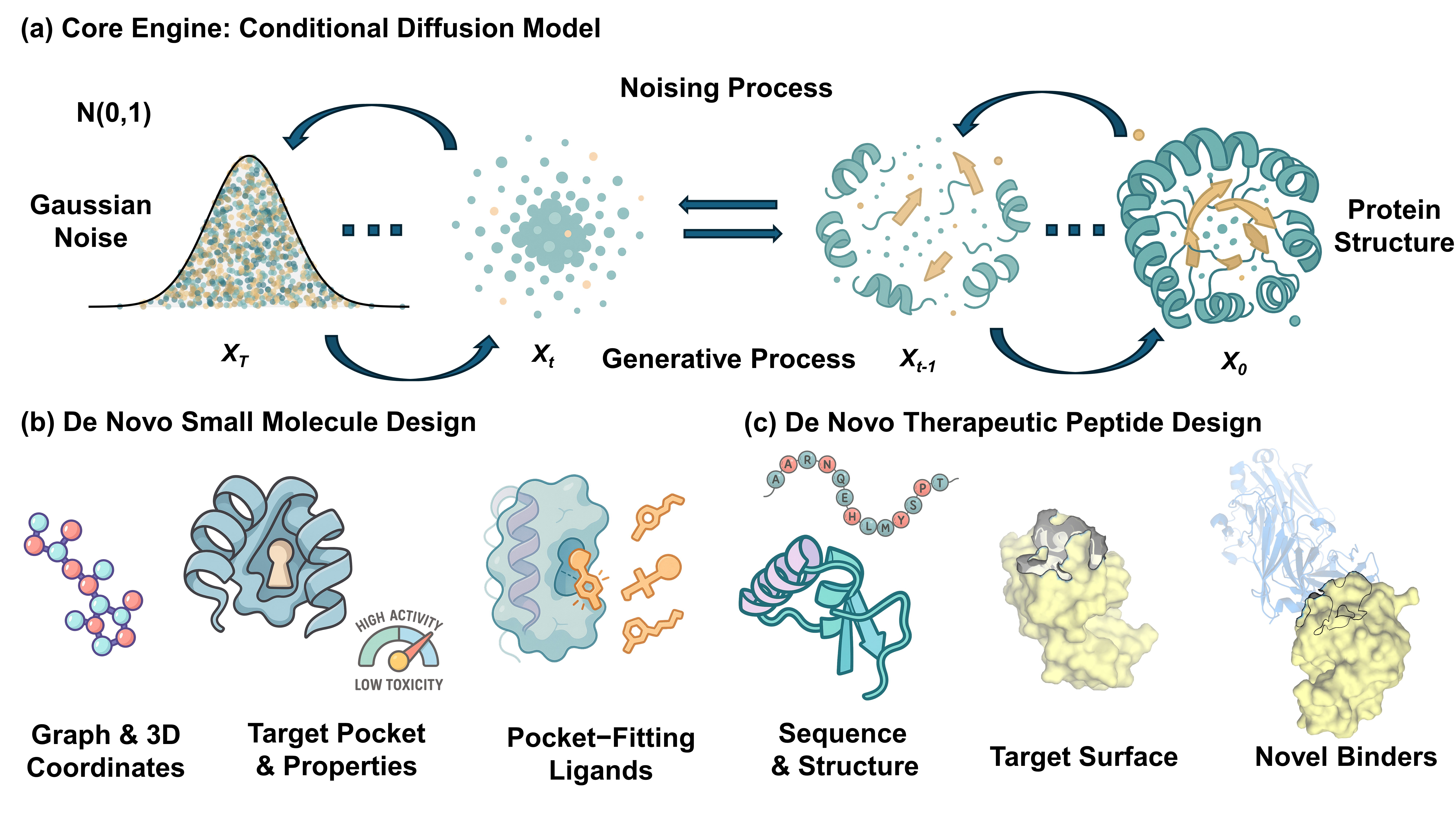}
		\caption{A unified framework for de novo drug design using a conditional diffusion model.
			(\textbf{a}) The core engine is a conditional diffusion model, which comprises two processes. The~\textit{noising process} systematically corrupts a data structure, such as a protein ($X_0$), into~Gaussian noise ($X_T$) over discrete timesteps. The~\textit{generative process} learns the reverse, creating novel structures by iteratively denoising from noise, guided by specific conditions.
			(\textbf{b}) For de novo small molecule design, the~model generates molecular graphs or 3D coordinates conditioned on a target's binding pocket and desired properties (e.g., high activity, low toxicity) to produce diverse, pocket-fitting ligands.
			(\textbf{c}) For de novo therapeutic peptide design, the~model generates peptide sequences and their corresponding 3D structures, conditioned on a target protein's surface, to~design novel binders.}
		\label{fig:framework}
	\end{figure*}

	In contrast, diffusion models circumvent many of these issues, which explains their recent ascendancy. Their training is stable and guided by a well-defined denoising objective, avoiding the adversarial instabilities of GANs while consistently producing samples of high quality and diversity~\cite{alakhdar2024diffusion, zhang2025unraveling, song2019generative, mullerfranzes2023multimodal, yang2023diffusion, wang2024gldm, brown2019guacamol}. Their framework is remarkably flexible, accommodating both continuous data like 3D coordinates and discrete data like atom types through tailored noise processes~\cite{hoogeboom2022equivariant, hu2025target, dunn2025flowmol3}. This adaptability, combined with powerful conditioning techniques like classifier-free guidance~\cite{ho2022classifier, zhang2024cross, yang2025diffmc}, allows for precise control over the iterative refinement process, leading to better global coherence and making them uniquely suited for the multifaceted challenges of molecular design. This entire process, from~the core diffusion engine to its specific applications in designing small molecules and therapeutic peptides, is conceptually illustrated in \cref{fig:framework}.
	
	These distinct generative mechanisms also lead to different trade-offs in computational cost, scalability, and~interpretability. While frameworks like VAEs and GANs typically employ their backbone in a single-pass, feed-forward manner for generation, diffusion models operate iteratively, requiring hundreds or thousands of denoising steps. This iterative paradigm makes them more computationally intensive at inference time. Although~the scalability of any single denoising step is governed by the underlying backbone (e.g., GNN or Transformer), the~total generation cost is this value multiplied by the number of iterations. This computational overhead, however, is often offset by superior training stability, as~diffusion models circumvent the notorious convergence issues and mode collapse that plague GANs. From~an interpretability perspective, all these deep generative models face the 'black box' challenge. Nevertheless, the~step-by-step refinement process of diffusion models may offer unique, albeit still nascent, opportunities for mechanistic insight by allowing observation of the generative trajectory, though~deciphering the rationale at each step remains an active area of research~\cite{ho2020denoising, saad2024survey, alakhdar2024diffusion, ho2022classifier}.

	\section{Application I: De Novo Design of Small~Molecules}\label{sec:small_molecules}
	\unskip
	
	\subsection{Datasets and Benchmarks for Small Molecule~Generation}
	The development and evaluation of diffusion models for small molecule design rely heavily on large-scale, high-quality datasets. The~most widely used benchmark is CrossDocked2020~\cite{Francoeur2020}, a~dataset containing approximately 22.5 million docked poses from over 100,000 protein-ligand complexes derived from the PDB (Protein Data Bank) through a systematic docking procedure~\cite{Berman2000}. Each complex includes the 3D coordinates of the protein binding pocket (typically defined as residues within 6--10 Å of the ligand) and the bound ligand, along with docking scores as a proxy for binding affinity. CrossDocked2020 has become the de facto standard for evaluating structure-based drug design models~\cite{corso2023diffdock,schneuing2024structure,hu2025target}, enabling direct comparison across different approaches including diffusion-based methods~\cite{alakhdar2024diffusion} and other generative AI techniques~\cite{das2025generative}. However, it has several acknowledged limitations: the docking scores are computational estimates rather than experimental measurements, the~dataset is biased toward certain protein families (kinases and proteases are over-represented), and~the ligands are primarily known drugs or drug-like molecules, limiting chemical diversity. These limitations have motivated ongoing efforts to develop more diverse and experimentally validated benchmarks for the~field.
	
	For property-based generation and conformer generation tasks, the~GEOM-Drugs dataset~\cite{axelrod2022geom} is commonly used, containing approximately 430,000 drug-like molecules with pre-computed 3D conformers generated using RDKit~\cite{landrum2013rdkit} and optimized with semi-empirical quantum chemistry methods. This dataset enables training of models that learn the distribution of molecular geometries and can generate diverse, low-energy conformers~\cite{zhu2022direct,morehead2024geometry,mcnutt2023conformer}. The~ZINC database~\cite{irwin2020zinc20}, containing over 230 million purchasable compounds, is often used for pre-training or as a source of negative examples. \textls[-10]{The~QM9 dataset~\cite{ramakrishnan2014quantum}, containing approximately 134,000 small organic molecules with quantum chemical properties computed at the DFT level, is used for benchmarking models on property prediction tasks, though~its molecules are smaller and simpler than typical drug~candidates.}
	
	A critical limitation across all datasets is the scarcity of experimentally validated binding affinity data~\cite{wei2025gatoraffinity, liu2024ppb, wang2024prediction}. While databases like BindingDB~\cite{liu2025bindingdb} and ChEMBL~\cite{gaulton2012chembl} contain millions of bioactivity measurements, only a small fraction include high-resolution 3D structures of protein-ligand complexes. Beyond~scarcity, the~quality and heterogeneity of this data present a fundamental challenge to model reproducibility. Bioactivity measurements (e.g., K\textsubscript{i}, K\textsubscript{d}, IC\textsubscript{50}) are often aggregated from diverse assays with varying experimental conditions, introducing significant noise and inconsistencies. This lack of standardized data curation and reporting protocols directly undermines a model's ability to learn robust structure-activity relationships, thereby compromising its generalization capabilities. Consequently, establishing rigorous data management standards is as critical as developing new algorithms for building truly predictive and reproducible generative models. This data challenge further motivates the development of transfer learning and semi-supervised approaches~\cite{krishnan2021accelerating, dalkiran2023transfer, buterez2024transfer, atz2024prospective, wang2025leveraging} that can leverage large unlabeled datasets while being robust to label~noise.
	
	\subsection{Structure-Based Drug Design (SBDD)}
	The central task in SBDD is to generate molecules that are geometrically and chemically complementary to a given protein binding pocket, maximizing binding affinity while maintaining drug-like properties. Diffusion models have shown remarkable success in this domain by learning to generate molecules directly in the 3D space of the binding~pocket.
	
	Pocket2Mol~\cite{peng2022pocket2mol}, one of the pioneering works in 2022, employs a two-stage approach: first generating molecular scaffolds as a set of 3D points, then predicting atom and bond types for these points. The~model is conditioned on pocket atom coordinates and features through a cross-attention mechanism, achieving 68.4\% pose selection accuracy on the CrossDocked2020 benchmark. The~model generates molecules with high validity ($>$95\%) and uniqueness ($>$90\%), demonstrating the capability of diffusion models to produce chemically valid~structures.
	
	DiffSBDD~\cite{schneuing2024structure} introduces an SE(3)-equivariant graph neural network architecture that jointly diffuses over atomic coordinates and discrete atom types. By~incorporating pocket information through a joint graph representation of the pocket and the growing molecule, DiffSBDD achieves superior performance in generating molecules with favorable predicted binding affinities. On~the CrossDocked2020 dataset, DiffSBDD generates molecules with a median Vina score of $-$7.5 kcal/mol. This score serves as a computational estimate of binding affinity, where more negative values indicate a stronger predicted interaction, and~this performance outperforms previous autoregressive and VAE-based approaches. Importantly, the~model demonstrates the ability to generate molecules that form key interactions (hydrogen bonds, hydrophobic contacts) with critical pocket residues, as~validated through molecular dynamics~simulations.
	
	TargetDiff~\cite{guan20233d,hu2025target} further advances the field by introducing a target-aware diffusion process that explicitly models the protein-ligand interaction energy during generation. \textls[-5]{By~incorporating a learned energy function that estimates binding affinity, TargetDiff demonstrates improved performance in generating high-affinity binders while maintaining molecular diversity across different regions of chemical space with strong pocket~complementarity.}
	
	Building upon these foundational approaches, recent work has explored dual diffusion frameworks and pharmacophore-oriented generation. Huang~et~al.~\cite{huang2024dual} introduced a dual diffusion model that enables both de novo 3D molecule generation and lead optimization, providing a unified framework for structure-based drug discovery. More recently, pharmacophore-oriented approaches~\cite{peng2025pharmacophore} have emerged to incorporate explicit constraints on the pharmacophore---the essential three-dimensional arrangement of molecular features required for biological activity---during the diffusion process, enabling more efficient feature-customized drug discovery by directly controlling key molecular properties and interaction~patterns.
	
	A primary challenge that remains is the precise modeling of key molecular interactions, such as hydrogen bonds, salt bridges, and~$\pi$-$\pi$ stacking~\cite{Zhung2024}. Furthermore, systematic benchmarks reveal persistent challenges in achieving accurate 3D spatial modeling, as~many generated structures show significant deviations from energy-minimized references, especially for larger molecules~\cite{qin2025comprehensive}. While current models can generate molecules that occupy the binding pocket, ensuring that specific pharmacophoric features are correctly positioned to form critical interactions with the protein remains difficult. Additionally, the~generated molecules often require post-processing steps, such as bond order correction and protonation state assignment, to~ensure chemical validity~\cite{Buttenschoen2024}.
	
	\subsection{Property-Based Ligand Design and~Optimization}
	This area focuses on generating molecules that satisfy multiple objectives simultaneously, such as high binding affinity, favorable drug-likeness measured by metrics like the Quantitative Estimate of Drug-likeness (QED)~\cite{Bickerton2012Quantifying}, where values closer to 1 suggest a better drug-like profile, appropriate lipophilicity (logP), low toxicity, high membrane permeability, and~synthetic accessibility (SA). To~achieve this multi-objective optimization, several property-guided generation approaches have been developed. Conditional diffusion models, for~example, are trained to generate molecules with specific property values by directly conditioning on target property vectors~\cite{hu2025target}. These models can produce molecules with specified molecular weight, logP, and~hydrogen bond donor/acceptor counts with reasonable accuracy~\cite{hu2025target}. Alternatively, guidance-based methods employ pre-trained property predictors to steer the diffusion sampling process at inference time~\cite{Oestreich2025DrugDiff,Han2023Training,alakhdar2024diffusion}. By~computing the gradients of property predictors with respect to the molecular representation, these techniques can navigate the chemical space to optimize multiple properties simultaneously~\cite{morehead2024geometry}.
	
	However, optimizing for multiple, often conflicting, objectives remains a significant challenge. For~instance, increasing lipophilicity (logP) to improve membrane permeability may concurrently decrease aqueous solubility and heighten toxicity risk. To~address this, recent work has explored more sophisticated frameworks. Some studies focus on generating diverse molecules along the Pareto front, providing a set of candidates that represent different trade-offs between objectives~\cite{KhodabandehYalabadi2025BoKDiff}. Other advanced strategies include using reinforcement learning to dynamically balance competing goals~\cite{Chen2025Uncertainty,Yuan2025A3D} and developing dual diffusion architectures for simultaneous optimization across multiple design criteria~\cite{yang2025diffmc}.
	
	However, ensuring the synthesizability of the generated molecules remains a major and persistent challenge in the field. While diffusion models can generate chemically valid molecules (as determined by valence rules and RDKit sanitization), these molecules may be synthetically inaccessible or require prohibitively complex synthetic routes. Synthetic accessibility scores (SA scores~\cite{ertl2009estimation}) provide a rough estimate~\cite{Oestreich2025DrugDiff,Yuan2025A3D}, but~they do not guarantee that a practical synthesis route exists. Recent efforts have focused on incorporating models for retrosynthesis, a~computational technique for planning chemical synthesis by working backward from the target molecule, into~the generation process, either by using retrosynthesis feasibility as an additional objective~\cite{guo2025directly} or by generating molecules in a retrosynthetically aware manner, building molecules from commercially available building blocks through known reaction templates~\cite{seo2023molecular,gainski2025scalable}. Alternative approaches evaluate synthesizability by combining retrosynthetic planning with forward reaction prediction to verify route feasibility~\cite{liu2024evaluating}. Methods that optimize molecular geometry and structural stability have also been proposed to improve the practical viability of generated candidates~\cite{morehead2024geometry}. Despite these advances, the~gap between computational generation and experimental synthesis remains a critical bottleneck~\cite{das2025generative,zeng2022deep}—a synthesis barrier that has been identified as a major challenge limiting the real-world impact of generative AI in pharmaceutical development~\cite{fu2025future}. Bridging this gap by integrating generative models with retrosynthesis prediction and automated experimental validation remains a central goal for the field~\cite{ramos2025review}, a~challenge shared across modalities, where the synthetic accessibility hurdle for small molecules finds its critical counterpart in the biological stability and production challenges inherent to therapeutic peptides (Section \ref{sec:peptides}).
	
	\section{Application II: Innovative Design of Therapeutic~Peptides}\label{sec:peptides}
	\unskip
	
	\subsection{Datasets and Benchmarks for Peptide~Design}
	Peptide and protein design models rely on fundamentally different datasets compared to small molecule models, reflecting the distinct nature of biopolymers. The~Protein Data Bank (PDB)~\cite{Berman2000}, containing over 240,000 experimentally determined protein structures (as of 2024), serves as the primary source of structural data. For~training diffusion models on protein backbones, high-quality subsets are typically used: the CATH database~\cite{orengo1997cath,sillitoe2015cath} (containing 601,493 domains from over 150,000 PDB structures, classified by architecture and topology) and the SCOPe database~\cite{fox2014scope,chandonia2019scope} (classifying 344,851 domains from 106,976~PDB entries by structural and evolutionary relationships) are commonly used to ensure structural diversity and avoid redundancy. These datasets enable models to learn the principles of protein folding—the allowed backbone geometries, secondary structure propensities, and~tertiary packing~arrangements.
	
	For sequence-based models, much larger datasets are available. UniProt~\cite{Apweiler2004,uniprot2025}, containing over 246 million protein sequences, provides a vast resource for learning sequence patterns and evolutionary relationships. The~UniRef50 and UniRef90 datasets~\cite{Suzek2007} (clustered at 50\% and 90\% sequence identity, respectively) are commonly used for training, providing non-redundant reference clusters that enable models to learn amino acid co-evolution patterns, functional motifs, and~sequence-structure relationships. The~recent AlphaFold Database~\cite{Varadi2022,Varadi2024}, containing predicted structures for over 214 million proteins, has dramatically expanded the available structure data, though~the quality varies and experimental validation is~limited.
	
	For specific peptide design tasks, specialized datasets exist. The~Antimicrobial Peptide Database (APD3) contains approximately 3000 experimentally validated antimicrobial peptides with activity data (MIC values, target organisms)~\cite{wang2016apd3}. The~Database of Antimicrobial Activity and Structure of Peptides (DBAASP) contains over 15,000 entries with detailed activity annotations~\cite{pirtskhalava2021dbaasp}. For~cell-penetrating peptides, CPPsite contains approximately 1800 entries~\cite{gautam2012cppsite, agrawal2016cppsite}. However, these specialized datasets are much smaller than those available for small molecules. Furthermore, they suffer from significant data heterogeneity, a~challenge that directly impacts model reproducibility and the creation of reliable benchmarks. For~instance, antimicrobial activity measured as Minimum Inhibitory Concentration (MIC) can vary dramatically depending on the bacterial strain and assay protocol, while cell-penetrating efficiency lacks a universally accepted standard metric. This inconsistency makes it difficult to harmonize data for training robust, generalizable predictive models and underscores the critical need for community-wide standards in peptide bioactivity~reporting.
	
	A critical challenge is the scarcity of experimentally validated peptide-protein interaction data with structural information. While databases like PDBbind~\cite{liu2015pdb,wang2005pdbbind} contain thousands of protein-ligand complexes, only a small fraction involve peptide ligands. The~lack of large-scale, high-quality training data for peptide binder design motivates the use of transfer learning from general protein structure prediction models (e.g., AlphaFold2~\cite{jumper2021highly}, RoseTTAFold~\cite{baek2021accurate}) and the development of physics-informed models that incorporate biophysical~priors.
	
	\subsection{Generation of Functional Peptide~Sequences}
	The goal here is to generate amino acid sequences with specific biological functions, such as antimicrobial peptides (AMPs), cell-penetrating peptides (CPPs), or~peptides with specific binding properties. This task typically employs discrete diffusion \mbox{models~\cite{alamdari2023protein,tang2025peptune,meshchaninov2024diffusion,luo2025cpldiff}}, which are adapted to handle the categorical nature of amino acid data. Pioneering work has demonstrated sequence-only generation without requiring structural information~\cite{alamdari2023protein}, with~recent advances enabling multi-objective optimization for therapeutic properties~\cite{tang2025peptune}, length-controlled peptide design~\cite{luo2025cpldiff}, and~applications in practical binder design~\cite{chen2025peptide,li2025thermodynamics}.
	
	Discrete diffusion models for peptide sequences operate by gradually corrupting amino acid sequences through a process of random token replacement or masking, then learning to reverse this process. The~foundational work in this area proposed several noise processes, including uniform transition matrices, absorbing state models, and~learned transition matrices that respect amino acid similarity~\cite{austin2021structured}. The~uniform transition approach, for~instance, has been applied in subsequent peptide generation models~\cite{alamdari2023protein}. The~choice among these noise processes carries significant practical implications for peptide design. The~uniform transition matrix, while the simplest to implement, disregards the inherent biochemical similarities between amino acids, treating a transition from Alanine to Valine (both hydrophobic) the same as one to Lysine (hydrophilic). The~absorbing state model is particularly well-suited for tasks like sequence inpainting or constrained generation, as~the \texttt{MASK} token provides a clear demarcation between fixed regions and those to be generated. Finally, learned transition matrices offer the most sophisticated approach, allowing the model to incorporate prior knowledge, such as amino acid substitution matrices (e.g., BLOSUM), which can potentially improve learning efficiency and generate more biologically plausible intermediate~sequences.
	
	Recent studies have demonstrated that deep generative and foundation models can successfully design antimicrobial peptides (AMPs) with predicted and experimentally validated activity comparable to, or~even exceeding, that of natural AMPs~\cite{Szymczak2023, Li2024, Dong2025, Wang2025}. Models are typically trained on curated datasets of a few thousand sequences drawn from larger public databases such as APD3, DBAASP, or~DRAMP, which contain up to  22,000~entries~\cite{Brizuela2025}. For~instance, a~recent generative model was trained on a specific set of 3280~MIC-labeled AMPs~\cite{Dong2025}. These approaches generate novel sequences with experimentally confirmed minimum inhibitory concentrations (MICs) in the low-micromolar range against common pathogens like \textit{E. coli} and \textit{S. aureus}; for example, validated MICs ranging from 0.20 to 15.18~$\upmu$M have been reported~\cite{Dong2025}, with~other generative frameworks also confirming potent hits~\cite{Szymczak2023}. Importantly, these generated peptides often exhibit substantial sequence novelty, with~one study reporting a median sequence identity of approximately 35\% to any example in the training set, indicating true \textit{de novo} design rather than memorization~\cite{Dong2025}.
	
	In peptide design, particularly for antimicrobial peptides (AMPs), diffusion models have been conditioned using strategies like text guidance or post-generation property filtering (e.g., net charge, hydrophobicity)~\cite{cao2025tg, jin2025ampgen}. The~application of similar methods for cell-penetrating peptides (CPPs), especially by explicit conditioning on predicted membrane permeability, is an emerging area that could leverage advances in CPP prediction models~\cite{seixas2024investigating}. Some generated peptides have demonstrated in~silico or in~vitro cellular uptake efficiencies comparable to canonical CPPs like TAT under specific assay conditions~\cite{tran2021using, gonzalez2023designing}, showcasing the potential to explore novel sequence space. However, systematic experimental validation remains a significant bottleneck. Recent reviews emphasize the persistent gap between computational predictions and functional confirmation, a~key challenge in translating in~silico designs into effective therapeutics~\cite{ramelot2023cell, lai2025deep, sutcliffe2024strategies}.
	
	A key advantage of diffusion models over previous generative approaches (such as RNNs or VAEs) is their ability to generate highly diverse sequences while maintaining exceptional validity~\cite{alamdari2023protein, zhang2023proldm, chen2024ampdiffusion, wang2025artificial}. Recent studies report that sequence validity—defined as the generation of valid amino acid strings of a desired length—consistently achieves near-perfect rates, typically $\ge$98--100\%~\cite{zhang2023proldm, wang2025artificial, alamdari2023protein}. Simultaneously, these models demonstrate substantially greater sequence diversity compared to VAE or language model baselines, producing broader and less redundant libraries that better span natural sequence and functional spaces~\cite{wang2025artificial, alamdari2023protein, zhang2023proldm}. While sequence-based generation is valuable for designing peptides with specific functional properties, many therapeutic applications require precise control over 3D structure and binding geometry. This motivates the development of structure-guided design approaches, which we explore~next.
	
	\subsection{Structure-Guided De Novo Peptide~Design}
	A more ambitious goal is to directly generate peptides that fold into specific 3D structures or bind to target protein surfaces with high affinity and specificity. This includes not only linear peptides but also larger, structurally defined mini-proteins that function as peptide mimetics. This task requires modeling both sequence and structure simultaneously, since the sequence must be compatible with the desired fold and the structure must be stable and functional~\cite{Rezaee2025Bridging,Wan2024Machine}. Recent deep learning advances, particularly diffusion-based methods, have made significant progress toward achieving this goal~\cite{Rettie2025Deep,Rettie2025Accurate}.
	
	RFdiffusion, a~landmark model in this area, has significantly advanced structure-guided protein and peptide design~\cite{watson2023denovo}. Built upon the RoseTTAFold structure prediction network~\cite{baek2021accurate}, RFdiffusion performs diffusion directly on protein backbone coordinates (represented as rigid body transformations of residue frames) while maintaining SE(3) equivariance~\cite{watson2023denovo}. The~model can be conditioned on various structural constraints, including target protein surfaces for binder design, desired secondary structure motifs (helices, sheets), or~functional site geometries~\cite{watson2023denovo}.
	
	RFdiffusion has demonstrated remarkable success in designing mini-protein binders, a~breakthrough that directly paves the way for creating structurally defined peptides with high efficacy~\cite{watson2023denovo}. When tasked with designing binders to challenging protein targets such as influenza hemagglutinin, for~instance, RFdiffusion generates backbones that, after~sequence design using ProteinMPNN~\cite{Dauparas2022}, achieve experimental binding affinities in the nanomolar range (e.g., a~K\textsubscript{D} of 28~nM for an influenza binder) in approximately 19\% of tested designs~\cite{watson2023denovo}. This success rate is substantially higher than previous computational design methods, which typically achieved success rates below 5\%~\cite{Cao2022, Bennett2023}. The~designed binders often exhibit novel folds not present in natural proteins, demonstrating the model's ability to explore diverse and novel structural topologies within the protein fold space~\cite{watson2023denovo}. Furthermore, the~approach has been successfully extended to designing high-affinity binders for challenging helical peptide targets, yielding picomolar to sub-nanomolar affinities~\cite{VazquezTorres2024}.
	
	The typical workflow, largely established by the developers of RFdiffusion~\cite{watson2023denovo}, is a critical hybrid approach involving two distinct generative stages. First, RFdiffusion (a diffusion model) is used to generate a peptide backbone (continuous coordinates) that is geometrically complementary to the target protein surface, with~the diffusion process conditioned on the target structure and desired binding interface residues. Second, a~sequence design model such as ProteinMPNN~\cite{Dauparas2022} (a GNN-based, non-diffusion model) or ESM-IF~\cite{lin2023evolutionary} is employed to perform inverse folding, designing an amino acid sequence (discrete tokens) compatible with the generated backbone. This two-step, hybrid methodology is significant because it highlights that structure-guided sequence design currently relies on integrating a powerful backbone DM with a specialized, non-diffusion inverse folding tool. A~pure diffusion model solution capable of generating both optimal structure and sequence simultaneously remains an active area of research. Third, the~resulting designs undergo computational validation using high-accuracy structure prediction models like AlphaFold2~\cite{jumper2021highly} or RoseTTAFold~\cite{baek2021accurate} to verify that the designed sequence folds into the intended structure and maintains the desired binding geometry. Finally, promising candidates proceed to experimental validation through protein expression, purification, and~binding~assays.
	
	Despite these successes, significant challenges unique to peptide therapeutics remain. Generated peptides must be engineered for proteolytic stability to overcome their inherently short in~vivo half-lives, a~consideration often addressed by incorporating non-canonical amino acids or cyclization, which are not yet fully integrated into diffusion workflows~\cite{Fetse2023}. Furthermore, minimizing potential immunogenicity by avoiding T-cell epitopes---specific peptide fragments recognized by the immune system that can trigger an unwanted response---is a critical design constraint that requires sophisticated predictive modeling~\cite{Achilleos2025}. Ultimately, ensuring that the designed sequence not only folds into the intended conformation but also remains stable and avoids aggregation is paramount, as~current models may not fully capture the subtle side-chain interactions governing these properties~\cite{Rettie2025}. Integrating these complex biological and biophysical constraints into the next generation of generative models represents a critical frontier for the~field.
	
	While the specific bottlenecks differ, the~parallel evolution of diffusion models in these two domains invites a systematic comparison. Synthesizing these distinct modality-specific insights is essential to identify the shared fundamental challenges and to envision the future trajectory of the field toward a unified, automated discovery~paradigm.
	\section{Comparison, Challenges, and~Future~Perspectives}\label{sec:comparison_future}
	\unskip
	
	\subsection{A Head-to-Head Comparison: Small Molecules vs.~Peptides}
\begin{figure*}[htb!]
		\includegraphics[width=\textwidth]{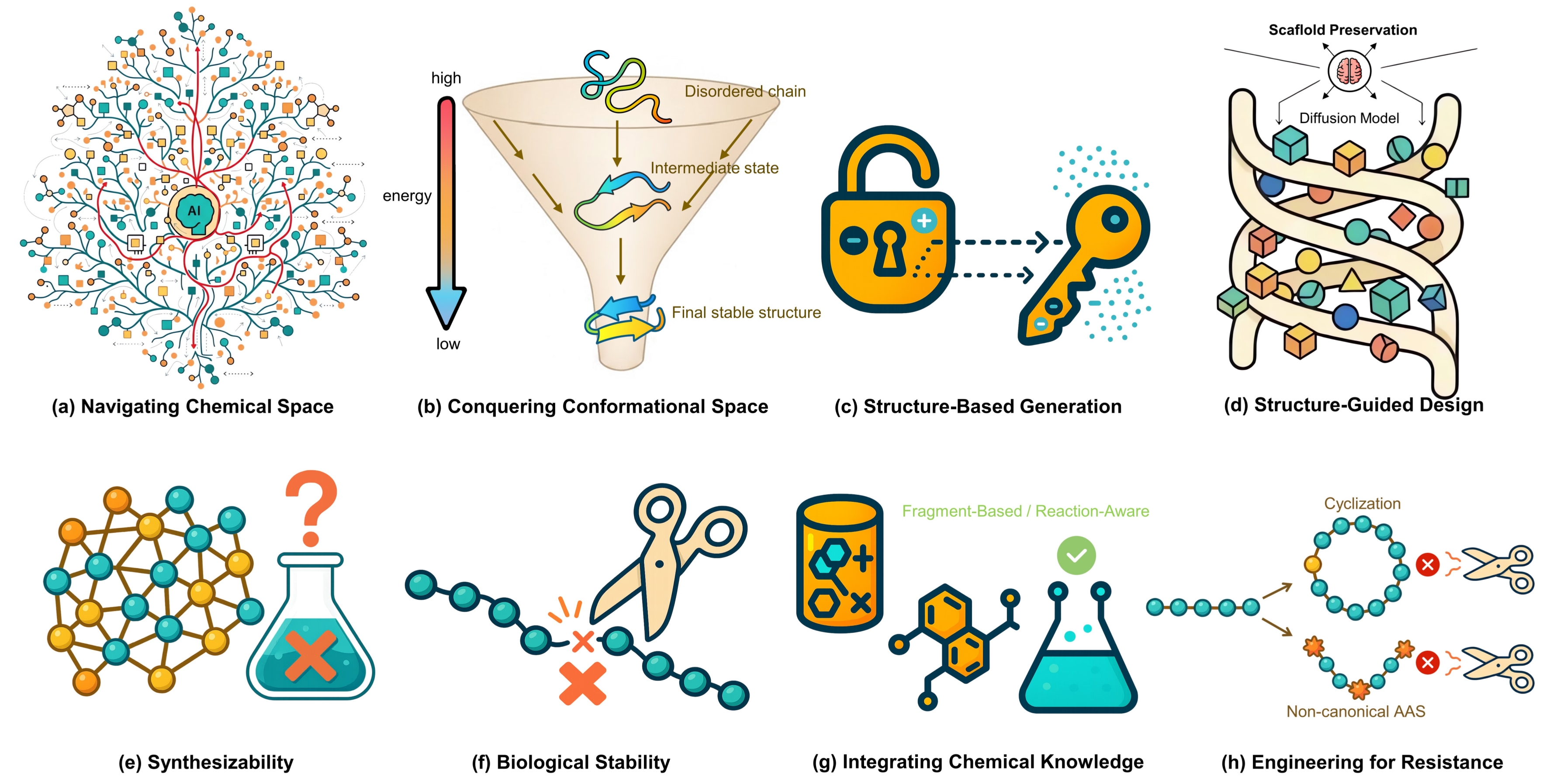}
		\caption{Contrasting Design Paradigms for Small Molecules and Therapeutic Peptides with Diffusion Models. 
			The figure illustrates the distinct challenges and tailored AI-driven solutions for small molecules (left column, \textbf{a},\textbf{c},\textbf{e},\textbf{g}) versus therapeutic peptides (right column, \textbf{b},\textbf{d},\textbf{f},\textbf{h}). 
			(\textbf{a},\textbf{b}) The primary challenge for small molecules is navigating the vast, discrete chemical space, whereas for peptides, it is conquering the continuous conformational space to achieve a stable fold. 
			(\textbf{c},\textbf{d}) Consequently, diffusion models are employed for structure-based generation to fit small molecules into binding pockets, while for peptides, they perform structure-guided design by decorating a predefined scaffold. 
			(\textbf{e},\textbf{f}) Key downstream hurdles also differ: ensuring chemical synthesizability for small molecules versus achieving biological stability against degradation for peptides. 
			(\textbf{g},\textbf{h}) Finally, solutions are modality-specific: integrating chemical knowledge (e.g., reaction rules) to guide synthesis for small molecules, and engineering stability in peptides through modifications like cyclization or using non-canonical amino acids. 
			Explanation of symbols: The red crosses (X) indicate synthetic infeasibility (e) or blocked enzymatic degradation (f, h). In (e), the colored spheres represent atoms within a complex molecular graph structure.}
		\label{fig:comparison}
	\end{figure*}

	The fundamental differences in applying diffusion models to small molecules and peptides are visually contrasted in \cref{fig:comparison} and further detailed in Table~\ref{tab:comparison}. This comparison highlights distinct challenges and opportunities in each domain, providing a clear framework for understanding the current landscape. As~illustrated, the~design of small molecules is fundamentally a challenge of navigating a vast, discrete chemical space to ensure chemical synthesizability, whereas peptide design is a problem of conquering a continuous conformational space to achieve biological stability. These core distinctions dictate everything from molecular representation to the primary validation hurdles, shaping two related yet distinct fields of AI-driven discovery. Beyond~these qualitative differences, quantitative performance metrics reveal the maturity and capabilities of current diffusion-based approaches in each domain, as~detailed in Table~\ref{tab:performance}.

	\begin{table*}[htb!]
		\caption{A Head-to-Head Comparison: Diffusion Models for Small Molecules vs.~Peptides.}
		\label{tab:comparison}
		
		\begin{adjustwidth}{-\extralength}{0cm}
			\begin{tabularx}{\fulllength}{@{} >{\bfseries}p{2.6cm} >{\RaggedRight}X >{\RaggedRight}X @{}}
				\toprule
				\textbf{Feature} & \textbf{Small Molecules} & \textbf{Therapeutic Peptides} \\
				\midrule
				\textbf{Representation} & 
				Graphs: Atoms \& bonds \newline 
				3D Point Clouds: Coordinates \newline 
				\textit{Requires $E(3)$ equivariance}~\cite{hoogeboom2022equivariant, morehead2024geometry, xu2022geodiff} & 
				Sequences: Discrete amino acids \newline
				3D Backbones: Continuous coordinates \newline
				\textit{Often requires distinct models for sequence (discrete) and structure (continuous) generation} \\\midrule
				
				\textbf{Chemical Space} & 
				Vast \& Discontinuous ($\sim$$10^{60}$)~\cite{Bohacek1996, Polishchuk2013, reymond2015chemical, Orsi2025} \newline
				Learns implicit chemical rules (e.g., valence) & 
				Combinatorial \& Structured ($20^n$)~\cite{Orsi2025} \newline
				Governed by protein folding~principles \\
				\midrule
				\textbf{Typical Size} & 
				MW: 150--900 Da (oral drugs often \mbox{300--500~Da)~\cite{doak2014oral}} \newline 
				Heavy Atoms: 10--50 \newline
				Mostly rigid structures & 
				MW: 500--5000 Da \newline
				Length: 5--50 amino acids~\cite{fosgerau2015peptide} \newline
				Highly flexible, multiple~conformations \\
				\midrule
				\textbf{Key Challenge} & 
				Synthesizability: Can it be made?~\cite{ertl2009estimation} \newline
				Stereochemistry control & 
				Biological Stability: Folding, proteolysis \newline
				Immunogenicity avoidance~\cite{fosgerau2015peptide} \\
				\midrule
				\textbf{Validation} & 
				Computational: Docking, ADMET~\cite{kitchen2004docking, daina2017swissadme} \newline
				Experimental: Synthesis, binding assays (SPR, ITC)~\cite{rich2000advances, myszka2000implementing, myszka1997kinetic} & 
				Computational: Structure prediction (AF2)~\cite{jumper2021highly} \newline
				Experimental: Expression, binding \& stability~assays \\
				\midrule
				\textbf{Conditioning} & 
				Protein pocket geometry~\cite{peng2022pocket2mol, guan20233d, schneuing2024structure} \newline 
				Pharmacophores, desired properties (QED, logP)~\cite{Bickerton2012Quantifying} &
				Target protein surface~\cite{watson2023denovo} \newline
				Structural motifs (helix), sequence~patterns \\
				\midrule
				\textbf{Data \& Cost} & 
				Data: PDBbind ($\sim$20k complexes), CrossDocked (~100k pairs) \newline 
				Cost: Varies widely by model and scale &
				Data: PDB ($\sim$220k entries), AlphaFold DB ($>$200 M structures) \newline
				Cost: Varies widely by model and~scale \\
				\midrule
				\textbf{Success Metrics} & 
				Chemical: Validity, Uniqueness, \mbox{Novelty~\cite{polykovskiy2020molecular, brown2019guacamol, schneuing2024structure}} \newline
				Predicted Affinity: High-affinity rate &
				Structural: Designability (folds to target)~\cite{jumper2021highly} \newline
				Experimental Success: Varies, often a few to tens of percent~\cite{watson2023denovo} \\
				\midrule
				\textbf{Example Works} & 
				Pocket2Mol~\cite{peng2022pocket2mol}, DiffSBDD~\cite{schneuing2024structure}, TargetDiff~\cite{guan20233d}, GeoDiff~\cite{xu2022geodiff}, DiffLinker~\cite{igashov2024equivariant} & 
				RFdiffusion~\cite{watson2023denovo}, ProteinMPNN~\cite{Dauparas2022} (seq. design), Chroma~\cite{ingraham2023illuminating}, EvoDiff~\cite{alamdari2023protein}, FoldingDiff~\cite{wu2024protein} \\
				\bottomrule
			\end{tabularx}
		\end{adjustwidth}
	\end{table*}
	\unskip

	\begin{table*}[htb!]
		\small
		\caption{Performance Highlights of Representative Models in Molecular~Generation.}
		\label{tab:performance}
		
		\begin{adjustwidth}{-\extralength}{0cm}
			\begin{tabularx}{\fulllength}{@{} >{\bfseries}p{2.8cm} >{\RaggedRight}p{4cm} >{\RaggedRight}X @{}}
				\toprule
				\textbf{Model} & \textbf{Modality/Role} & \textbf{Key Performance Metrics \& Highlights} \\
				\midrule
				\multicolumn{3}{@{}l}{\textit{\textbf{Small Molecule Generation (Diffusion Models)}}} \\ 
				\midrule
				\textbf{Pocket2Mol} \cite{peng2022pocket2mol} & Structure-based generation & Avg. Vina score: $-$7.29 kcal/mol; High-affinity rate: 54.2\%; Good drug-likeness (QED: 0.56). \\
				\textbf{DiffSBDD} \cite{schneuing2024structure} & Structure-based generation & High chemical validity (97.8\%) and novelty (85.7\%); Median Vina score: $-$7.50 kcal/mol. \\
				\textbf{TargetDiff} \cite{guan20233d} & Guided generation & State-of-the-art binding affinity (Avg. Vina: $-$7.80 kcal/mol); High-affinity rate: 58.1\%. \\
				\textbf{GeoDiff} \cite{xu2022geodiff} & Conformer generation & High-quality 3D conformer generation with low geometric error (MAT-R: 0.86 Å on Drugs dataset). \\
				\midrule
				\multicolumn{3}{@{}l}{\textit{\textbf{Peptide and Protein Design (Diffusion-Centric Workflows)}}} \\
				\midrule
				\textbf{RFdiffusion} \cite{watson2023denovo} & Backbone generation (Diffusion) & High experimental success rate for binders (14--19\%); Generated structures match Cryo-EM to 0.63 Å RMSD. \\
				\textbf{ProteinMPNN} \cite{Dauparas2022} & Sequence design (GNN, non-diffusion) & High native sequence recovery (52.4\%); Essential downstream tool for designing sequences for generated~backbones. \\
				\textbf{Chroma} \cite{ingraham2023illuminating} & Protein/Complex generation (Diffusion) & Experimentally confirmed designs with crystal structures matching to \textasciitilde1.0 Å RMSD; Generates diverse~topologies. \\
				\textbf{EvoDiff} \cite{alamdari2023protein} & Sequence generation (Discrete Diffusion) & High experimental success for functional proteins (65--75\%); Generates evolutionarily plausible~sequences. \\
				\bottomrule
			\end{tabularx}
		\end{adjustwidth}
	\end{table*}
	\unskip
	
	\subsection{Shared Hurdles and Common~Challenges}
	Despite their fundamental differences, the~deployment of diffusion models in both small molecule and peptide design is hampered by several shared, fundamental obstacles. Perhaps the most universal bottleneck is the reliance on imperfect scoring functions to evaluate generated candidates. Current approaches depend heavily on computational proxies like docking scores or predicted affinities, which often show poor correlation with experimental reality and lead to high false-positive rates in downstream \mbox{validation~\cite{trott2010autodock, eberhardt2021autodock, friesner2004glide, hollingsworth2018molecular, hospital2015molecular, jimenez2018k, wang2015accurate, kwon2020ak-score, lee2024improved}}. This challenge is directly exacerbated by the scarcity of high-quality labeled data. While vast repositories exist~\cite{kim2023pubchem, kim2016pubchem, irwin2020zinc20, Berman2000, Varadi2022, Varadi2024}, data that pairs molecular structures with experimentally validated, high-fidelity biological activity or binding affinity is a rare commodity, limiting the predictive power of supervised models~\cite{Francoeur2020, Wu2024Deep}. Promising mitigation strategies include physics-informed modeling, active learning, and~transfer learning, but~fundamental limitations remain~\cite{bhati2021pandemic, filella-merce2025optimizing, gorantla2024benchmarking, bailey2024deep, loeffler2024optimal, Goles2024Peptide, Wan2024Machine, Rezaee2025Bridging, AlOmari2024Accelerating}.
	
\begin{figure*}[htb!]
		\includegraphics[width=1\textwidth]{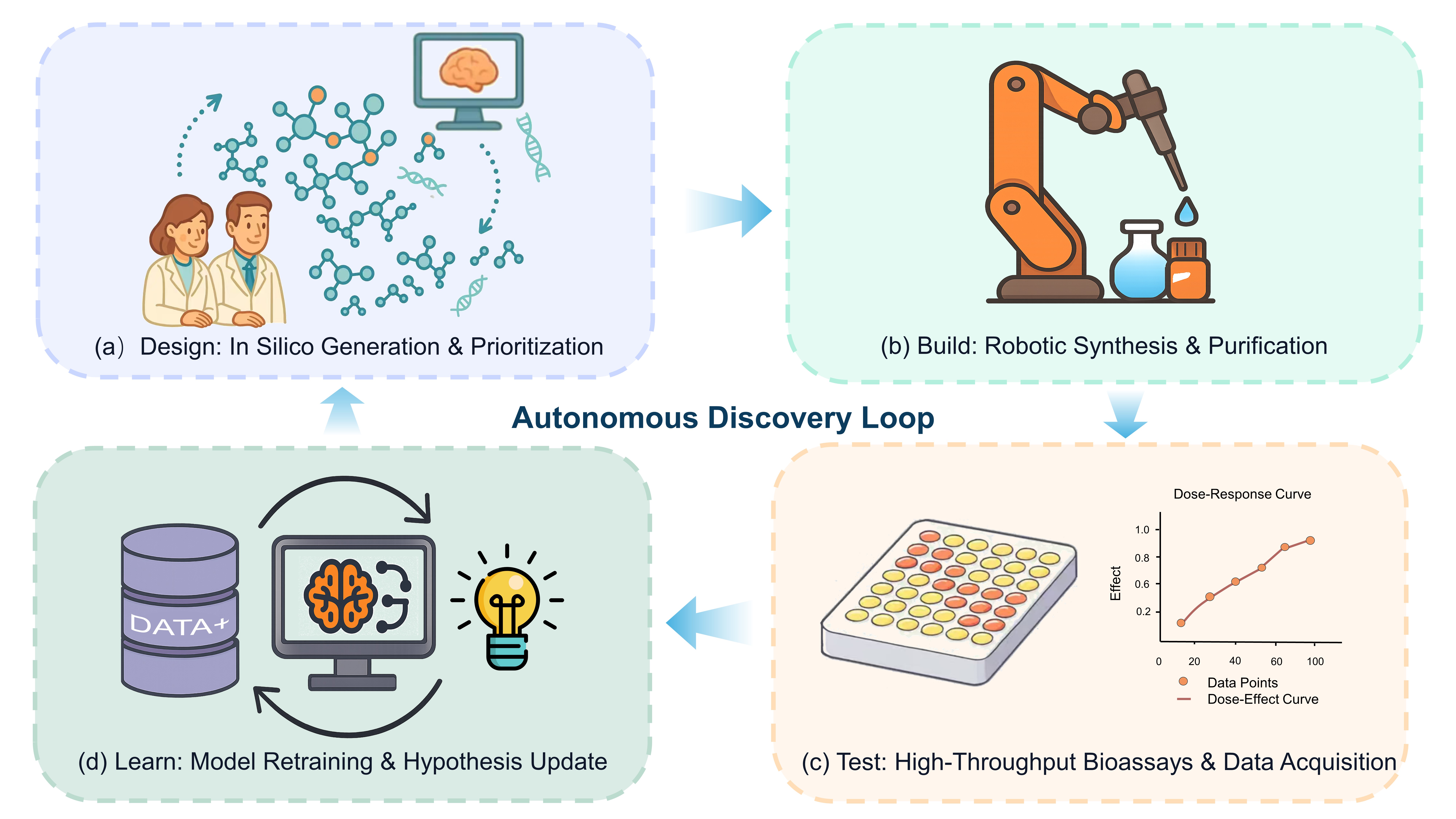}
		\caption{A Closed-Loop Paradigm for Drug Discovery Driven by AI and Automation. The~figure depicts an autonomous Design-Build-Test-Learn (DBTL) cycle, representing a future paradigm for accelerated therapeutic discovery. This approach seamlessly integrates AI-powered design with automated laboratory execution to create a self-optimizing discovery engine. 
			(\textbf{a}) Design: Generative AI models propose novel molecular candidates in~silico. 
			(\textbf{b}) Build: The most promising candidates are synthesized and purified using robotic platforms. 
			(\textbf{c}) Test: The synthesized compounds are evaluated in high-throughput biological assays to generate activity data. 
			(\textbf{d}) Learn: Experimental results are fed back into the AI model, which updates its knowledge and generates more informed hypotheses for the next cycle. This iterative process aims to dramatically shorten timelines and increase the success rate of finding novel medicines.} 
		\label{fig:loop}
	\end{figure*}

	Consequently, a~critical imperative for the field is to ``close the loop'' by integrating generative models with automated experimental validation in a Design-Build-Test-Learn (DBTL) cycle, as~illustrated in \cref{fig:loop}~\cite{Matzko2024Technologies,NAS2025Age,Liao2022Artificial}. The~implementation of such a cycle creates a direct pathway from in~silico hypothesis generation to experimental validation and back, enabling a rapid, iterative flow where data from one round directly informs the next. Without~such a framework, which is now becoming feasible through advances in laboratory automation~\cite{Abolhasani2023Rise,Dai2024Autonomous,Tom2024Self,Ha2023AI}, the~design process remains a slow, sequential, and~inefficient endeavor~\cite{Kusne2020Onthefly,Ramos2023Bayesian,Xian2025Unlocking,Wu2024Race}. Finally, even with better data and validation, the~issue of generalization persists. Like all machine learning models, diffusion models risk overfitting to their training distribution, potentially failing to generate effective and novel solutions for new biological targets or chemical spaces that lie outside their learned domain~\cite{Klarner2024Contextguided, Oestreich2025DrugDiff, Wacker2017How, Chen2024Design}. Overcoming these interconnected challenges is essential to translate the theoretical promise of diffusion models into tangible therapeutic breakthroughs~\cite{Khoee2024Domain, Xie2025Accelerating}.
	
	\unskip
	
	\subsection{Future Outlook and~Opportunities}
	A critical measure of success for any drug discovery technology is its path to clinical translation. While molecules designed specifically by diffusion models have not yet entered human clinical trials, the~broader field of generative AI provides a strong and encouraging precedent. Companies such as Insilico Medicine, Exscientia, and~Recursion Pharmaceuticals have successfully advanced AI-designed small molecules into various stages of clinical development, validating the principle that AI can indeed yield viable therapeutic candidates~\cite{Dharmasivam2025Leading, Dermawan2025From}. This industry progress sets the stage for diffusion models, which represent the state-of-the-art in generative capability. Several leading biotechnology companies are now deeply integrating these models into their R\&D pipelines, and~although much of this work remains proprietary, early reports indicate that diffusion-generated candidates are demonstrating excellent activity and favorable properties in preclinical studies. This marks a rapid transition of the technology from academic exploration to industrial~application.

	The field of diffusion models for drug discovery is rapidly evolving, with~future work poised to address current limitations and unlock transformative capabilities. A~key frontier is the development of unified frameworks—so-called ``foundation models'' for molecular science—that could seamlessly design not only small molecules and peptides but also complex hybrid therapeutics like peptide-drug conjugates (PDCs) and PROTACs from a single, powerful architecture. Enhancing model reliability is also paramount; this involves a shift from `black box' generators to interpretable and controllable tools that empower expert-guided design, while integrating first-principles simulations from quantum chemistry and physics to ensure the physical realism of generated candidates. Ultimately, the~successful translation of these technologies will hinge on fully realizing the automated Design-Build-Test-Learn (DBTL) paradigm, as~illustrated in \cref{fig:loop}, which promises to accelerate discovery cycles from months to days. This acceleration, however, must be navigated alongside the establishment of clear ethical and regulatory frameworks to guide AI-designed therapeutics safely from concept to~clinic.
	
	\section{Conclusions}
	Diffusion models have emerged as a powerful, unified generative framework, demonstrating remarkable versatility in designing both small molecules and therapeutic peptides. While successful in generating novel candidates for both modalities, the~path to clinical translation is defined by distinct, fundamental hurdles: for small molecules, the~challenge lies in bridging the gap from computational validity to practical chemical synthesizability; for peptides, it is ensuring that \textit{de novo} structural designs achieve in~vivo biological stability and function. Crucially, the~progress of AI-designed drugs now entering clinical trials provides a strong tailwind for the field, validating the potential of these advanced generative approaches. The~full potential of this technology will be significantly accelerated by closing the Design-Build-Test-Learn loop through deep integration with laboratory automation, which will enable rapid, data-driven iteration. By~overcoming these challenges, diffusion models hold the promise to catalyze a fundamental shift in drug discovery—moving from the passive exploration of existing chemical space to the active, purpose-driven creation of novel~medicines.
	
	\section*{Funding}
	This work was supported by the Natural Science Foundation of Xinjiang Uygur Autonomous Region (Grant Number: 2024D01C216) and the ``Tianchi Talents'' introduction plan.
	
	\section*{Author Contributions}
	Y.W. and Y.M. contributed equally to this work. Conceptualization, K.W.; Investigation, Y.W. and Y.M.; Writing—Original Draft, Y.W., Y.M., Y.C., J.Y., J.Z., M.C. and K.W.; Visualization, Y.W., Y.C., J.Y., J.Z. and M.C.; Writing—Review \& Editing, all authors; Supervision and Project Administration, K.W. All authors have read and agreed to the published version of the manuscript.
	
	\section*{Conflicts of Interest}
	The authors declare no conflict of interest.
	
	\bibliographystyle{unsrt} 
	\bibliography{references}
\end{document}